\documentclass[letterpaper, 10 pt, conference]{ieeeconf}
\usepackage[dvipsnames]{xcolor} 
\usepackage{algorithm}
\usepackage{algpseudocode}
\usepackage{algorithmicx}

\usepackage{array}
\usepackage{textcomp}
\usepackage{stfloats}
\usepackage{changes}
\usepackage{subcaption}
\usepackage{url}
\usepackage{verbatim}
\usepackage{graphicx}
\usepackage{cite}
\usepackage{hyperref}

\usepackage{amsmath, amsfonts, amssymb, amsthm}
\usepackage{mathtools}
\usepackage{ulem}
\usepackage{booktabs}

\usepackage{siunitx}    

\usepackage[T1]{fontenc} 

\IEEEoverridecommandlockouts
\hypersetup{hidelinks}
\usepackage{multirow}
\usepackage{multicol}
\usepackage[table]{xcolor}
\begin{document}
\title{\LARGE \bf Rethinking Reference Trajectories in Agile Drone Racing:\\A Unified Reference-Free Model-Based Controller via MPPI}
\author{Fangguo Zhao$^{1}$, Xin Guan$^{1}$, and Shuo Li$^{1}$
\thanks{$^{1}$Authors are with the College of Control Science and Engineering, Zhejiang University, Hangzhou 310027, China {\tt\small shuo.li@zju.edu.cn}}%
}
\maketitle
\begin{abstract}
While model-based controllers have demonstrated remarkable performance in autonomous drone racing, their performance is often constrained by the reliance on pre-computed reference trajectories. Conventional approaches, such as \textit{trajectory tracking}, demand a dynamically feasible, full-state reference, whereas \textit{contouring control} relaxes this requirement to a geometric path but still necessitates a reference. Recent advancements in reinforcement learning (RL) have revealed that many model-based controllers optimize surrogate objectives, such as trajectory tracking, rather than the primary racing goal of directly maximizing progress through gates. Inspired by these findings, this work introduces a reference-free method for time-optimal racing by incorporating this gate progress objective, derived from RL reward shaping, directly into the Model Predictive Path Integral (MPPI) formulation. The sampling-based nature of MPPI makes it uniquely capable of optimizing the discontinuous and non-differentiable objective in real-time. We also establish a unified framework that leverages MPPI to systematically and fairly compare three distinct objective functions with a consistent dynamics model and parameter set: classical trajectory tracking, contouring control, and the proposed gate progress objective. We compare the performance of these three objectives when solved via both MPPI and a traditional gradient-based solver. Our results demonstrate that the proposed reference-free approach achieves competitive racing performance, rivaling or exceeding reference-based methods. Videos are available at \url{https://zhaofangguo.github.io/racing_mppi/}

\end{abstract}

\section{Introduction}

In recent years, autonomous quadrotors have consistently broken speed records and demonstrated numerous potential applications in unknown areas \cite{10530312}. One of the driving forces behind these new records is autonomous drone racing, a platform for testing various elements of a drone's autonomous agile flight, including navigation, trajectory generation, and control techniques. For the aggressive and near time-optimal flight, a traditional feedback controller may not provide precise tracking. Model-based method shows great potential for time-optimal flight\cite{foehn2021time,9794477,romero2022model}, but relies on a precomputed reference trajectory. In this work, we focus on this problem and propose a method that can achieve a near time-optimal flight without a reference trajectory.

A prevailing paradigm in autonomous drone racing is the \textit{plan-then-track} framework, which first generates a high-performance trajectory and subsequently employs a model-based controller for tracking. Early successes in this area were built upon differential flatness and minimum-snap techniques \cite{mellinger2011minimum, faessler2017differential}, with numerous follow-up works enhancing computational efficiency and flight safety \cite{WANG2022GCOPTER, 9543598, 10342456, fork2023euclidean}. Recently, MINCO was developed to handle complex constraints and improve the solve efficiency\cite{WANG2022GCOPTER,qin2024timeoptimalplanninglongrangequadrotor}. Some work uses the Complementary Progress Constraint (CPC) method\cite{foehn2021time,10341844}, which treats the trajectory generation as a discrete nonlinear optimization problem capable of generating a global time-optimal reference trajectory. A common thread in these works is their heavy reliance on a complete, time-parameterized, full-state reference trajectory, which is then tracked by controllers such as a geometric controller or a Nonlinear Model Predictive Control (NMPC) scheme. To generate such references, these methods typically face a trade-off: they either compute the trajectory offline to guarantee time-optimality, or they replan online at the cost of sacrificing this optimality guarantee.

A notable shift towards reducing this dependency is marked by the development of Model Predictive Contouring Control (MPCC) \cite{romero2022model,romero2022time,krinner2024mpccmodelpredictivecontouring,11128227,11128315}. Instead of rigidly tracking a state at a specific time, MPCC maximizes progress along a geometric path while minimizing tracking error. This significantly relaxes the requirements on the reference, as it only necessitates an arc-length parameterized path defined by spatial positions, rather than a full-state, dynamically feasible trajectory. This simplified, position-only reference is computationally tractable enough to be generated online using simplified dynamics like point-mass models or through learning-based approaches \cite{11128227,11128315}. This represents a crucial step in decoupling the controller from strict temporal constraints. However, these methods still depend on a reference path, which serves as a necessary geometric prior.

More recently, the advent of reinforcement learning (RL) has introduced a new paradigm in autonomous drone racing, eliminating the need for a reference trajectory altogether \cite{song2021autonomous,song2023reaching,kaufmann2023champion}. These state-of-the-art methods are trained end-to-end for specific tracks, optimizing a \textit{gate progress} objective that directly rewards advancement along the race course. While highly effective, this performance is achieved by sampling and interacting with the environment extensively, causing the policy to overfit to a specific track and thereby limiting its generalization capabilities. Recent works treat the environment as a policy to adapt the control policy to new tracks, but need to train another network \cite{wang2025environmentpolicylearningrace}. Conversely, model-based controllers employing a Real-Time Iteration (RTI) scheme inherently possess strong generalization. This contrast presents a critical research question: how can the superior, task-centric objective from RL be integrated with the generalizability of model-based control? A fundamental challenge hinders this integration. The \textit{gate progress} objective, by its nature, is discontinuous and non-differentiable\cite{song2023reaching}. While some work has incorporated gate-relative terms into MPC formulations, they often serve only as a heuristic component rather than the primary driving objective, thus failing to fully leverage the potential of a gate-oriented strategy \cite{11127454}. This characteristic renders it incompatible with the gradient-based optimization solvers that are foundational to traditional model-based frameworks like Nonlinear Model Predictive Control (NMPC) and Model Predictive Contouring Control (MPCC).

In this work, we leverage the sampling-based nature of the Model Predictive Path Integral controller to directly integrate the RL-inspired gate progress objective into a model-based control paradigm. This yields a reference-free method that navigates a race track based solely on the waypoints themselves. Simultaneously, the MPPI framework is uniquely suited not only to handle such non-differentiable, task-centric objectives but also to solve traditional objectives for MPC and MPCC. Capitalizing on this versatility, we establish a unified and fair testbed. By employing an identical dynamics model, a single solution framework, and a consistent system input structure, we conduct the first direct comparison of three distinct objective functions under the same conditions. Our framework is designed with high modularity, allowing different control objectives to be interchanged with minimal code modification. 

The contributions of our study are as follows:
\begin{enumerate}
    \item A novel reference-free method for drone racing that leverages a gate progress objective within an MPPI framework.

    \item A unified control framework based on MPPI. Establishing a fair and consistent testbed for direct performance comparison.

    \item Simulations and real-world experiments validate that the proposed reference-free \textit{gate progress} objective achieves near time-optimal performance, and our MPPI-based implementations of the traditional objectives achieve highly competitive performance against their gradient-based counterparts, surpassing them in various simulated scenarios.

\end{enumerate}

\section{Related Work}

\subsection{Derivative-Free Model Predictive Control}
A significant limitation of traditional gradient-based solvers is their requirement for the objective function to be continuous and differentiable. To overcome this, a distinct line of research employs sampling-based, derivative-free optimization to navigate complex cost landscapes. Among these methods, Model Predictive Path Integral (MPPI) control \cite{williams2018information} has emerged as a particularly prominent and effective approach. MPPI leverages Monte Carlo sampling to iteratively refine a control sequence, making it inherently capable of handling discontinuous and non-differentiable objective functions that are intractable for gradient-based solvers. The efficacy of MPPI has been significantly amplified by modern parallel computing architectures, enabling massive-scale sampling and evaluation on GPUs to achieve real-time performance \cite{minavrik2024model}. This has led to its successful application in a wide array of challenging robotics problems, including contact-rich scenarios \cite{xue2024full}, tracking of dynamically infeasible trajectories \cite{yi2024covo,huang2023datt}, and planning under environmental uncertainty \cite{higgins2023model}. The highly parallelizable nature of the MPPI algorithm makes it a promising candidate for deployment on embedded systems, particularly as parallel computing capabilities on edge devices continue to advance.

\subsection{Comparative Analysis of Drone Racing Controllers}
The proliferation of diverse controller designs in autonomous drone racing has spurred a critical line of research aimed at identifying the optimal approach for a given task. Comparative studies have generally proceeded along two main fronts: fine-tuning components within a single control paradigm and benchmarking disparate control methodologies. The former category includes meticulous comparisons of design choices for reinforcement learning (RL) agents, such as the impact of different action spaces \cite{kaufmann2022benchmark} or policy input features \cite{dionigi2024power}. The latter, broader category encompasses comparisons across different classes of controllers, such as those between NMPC and differential flatness-based methods \cite{9794477} or between RL and geometric controllers for trajectory tracking \cite{kunapuli2025leveling}. While some studies have utilized MPPI, demonstrating its superior tracking performance over MPC \cite{minavrik2024model} or benchmarking it against gradient-based solvers for similar problems \cite{app15169114}, its potential as a unifying framework has been largely overlooked. The capacity of MPPI to accommodate disparate objective functions—bridging the gap between traditional and task-centric formulations—remains a largely unexplored avenue for enabling truly fair and systematic controller comparisons.

\section{PRELIMINARY}
\begin{figure*}
    \centering
    \includegraphics[width=1.0\linewidth,trim={0 85cm 0cm 0cm}, clip]{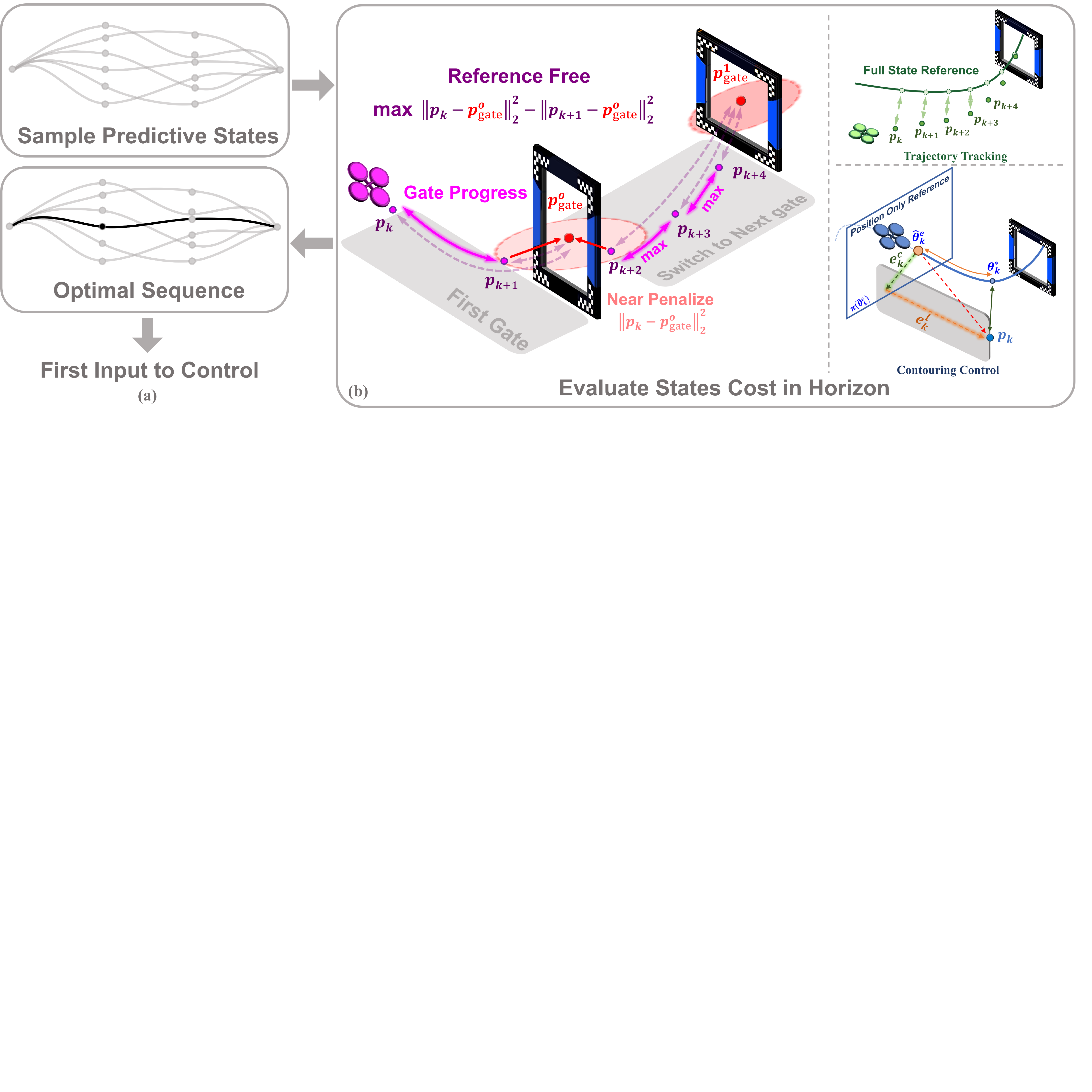}
    \caption{A conceptual illustration of the MPPI framework. (a) Sample and weighted stage of the MPPI. (b) Proposed reference-free gate progress objective with two reference-based objectives: Trajectory Tracking and Contouring Control.}
    \label{fig:three_objective}
    \vspace{-1em}
\end{figure*}
To solve the time-optimal drone racing problem, a common paradigm is to first generate a kinematically feasible, aggressive reference trajectory and then employ a controller to track it. Due to the highly aggressive nature of such trajectories, which often push the vehicle to its physical limits, traditional feedback controllers struggle to maintain stability and tracking performance. Consequently, model-based optimal control methods have become prevalent. Model-based controllers leverage an internal dynamics model to predict the system's future evolution and optimize the control inputs over a finite horizon, making it highly effective for such demanding tasks. The general formulation of a model-based controller is:
\begin{align}
\min_{\mathbf{U}} \quad&\sum_{k=0}^{K-1} C(\mathbf{x}_k,\mathbf{u}_k) + C(\mathbf{x}_K) \label{equ:MPC}\\
\textit{s.t.}\quad&\mathbf{x}_{k+1}=f(\mathbf{x}_k, \mathbf{u}_k), \quad \mathbf{x}_0=\mathbf{x}_{\text{init}} \nonumber
\end{align}
where $\mathbf{x}_k$ and $\mathbf{u}_k$ are the system state and control input at prediction step $k$, respectively, and $f(\cdot)$ is the discrete-time dynamics model with time step $\Delta t$. The objective is to find an optimal control sequence $\mathbf{U}^*=\{\mathbf{u}_0^*, \dots, \mathbf{u}_{K-1}^*\}$ that minimizes the sum of a stage cost $C(\mathbf{x}_k, \mathbf{u}_k)$ and a terminal cost $C(\mathbf{x}_K)$. In accordance with the receding horizon principle, only the first input $\mathbf{u}_0^*$ is applied to the system. The optimization problem is then resolved at the subsequent time step, a process often accelerated by schemes like the Real-Time Iteration (RTI).

In this section, we first present the quadrotor dynamics model, followed by an introduction to the Model Predictive Path Integral (MPPI) control algorithm, which we employ as our optimizer.

\subsection{Quadrotor Dynamics Model}
The model-based controllers utilize the dynamic functions of the quadrotors to predict future states using control inputs. The dynamic function $ f(\boldsymbol{x}_k, \boldsymbol{u}_k)$ represents the quadrotor's dynamics model. For clarity, we provide the dynamics model below:
\begin{align}
\prescript{}{}{\dot{\mathbf{x}}} = \mathbf{f}_{dyn}(\prescript{}{}{\mathbf{x}},\prescript{}{}{\mathbf{u}}) =
\begin{cases}
\prescript{}{}{\mathbf{v}} \\
\mathbf{g}+\frac{1}{\prescript{}{}{m}}\mathbf{R}(\prescript{}{}{\mathbf{q}}){\prescript{}{}{\mathbf{T}}} - \mathbf{R}(\mathbf{q}) \mathbf{D} \mathbf{R}^T(\mathbf{q}) \cdot \mathbf{v} \\
\frac{1}{2}\Lambda(\prescript{}{}{\mathbf{q}})
\begin{bmatrix}
0 \\ \prescript{}{}{\boldsymbol{\omega}} 
\end{bmatrix}\\
\mathbf{J}^{-1}(\boldsymbol{\tau}- \boldsymbol{\omega}\times\mathbf{J}\boldsymbol{\omega})
\end{cases},
\label{equ:dynamics model}
\end{align}
where 
\begin{align*}
    {\prescript{}{}{\mathbf{T}}} = \begin{bmatrix}
        0 \\  0 \\ \sum \prescript{}{}{{T}_s}
    \end{bmatrix},
    \boldsymbol{\tau} = 
\begin{bmatrix}
\frac{l}{\sqrt{2}} (T_1 + T_2 - T_3 - T_4) \\
\frac{l}{\sqrt{2}} (-T_1 + T_2 + T_3 -T_4) \\
c_{\tau}(T _1 - T_2 + T_3 - T_4)
\end{bmatrix},
\end{align*}
where, $\boldsymbol{T} $ is the thrust vector, and $\boldsymbol{\tau} $ is the torque vector. The variable $\prescript{}{}{{T}_s} $ represents the thrust generated by the rotors. In the above equations, $\prescript{}{}{\mathbf{x}} = [\prescript{}{}{\boldsymbol{p}}, \prescript{}{}{\boldsymbol{v}}, \prescript{}{}{\boldsymbol{q}}, \prescript{}{}{\boldsymbol{\omega}}] $ denotes the position, velocity, quaternion, and angular velocity of the quadrotor, respectively. The rotation matrix is represented by $\mathbf{R}(\prescript{}{}{\mathbf{q}}) $. Following the approach in \cite{romero2022model}, we define the control input $\mathbf{u}$ as the derivative of the single rotor thrust, i.e., $\mathbf {u} = \dot{\mathbf{T}}=[\dot{T}_1, \dot{T}_2, \dot{T}_3, \dot{T}_4]$.

\subsection{Model Predictive Path Integral Control Framework}
To solve the general optimal control problem presented in (\ref{equ:MPC}), we employ the Model Predictive Path Integral (MPPI) framework. MPPI is a sampling-based algorithm that leverages Monte Carlo methods to approximate the solution, thereby avoiding the explicit gradient computations required by traditional trajectory optimization frameworks. This key feature makes it particularly well-suited for objectives that may be complex or non-differentiable.

The optimization process at each time step begins by generating $M$ candidate control sequences, $\mathbf{U}^{m}=[u_0^m,u_1^m...u_{K-1}^m]$, by perturbing the optimal sequence from the previous iteration, $\mathbf{U^*}$:
\begin{equation}
    \mathbf{U}^{m} = \mathbf{U}^* + \delta\mathbf{U}^{m},
    \label{eq:mppi_sampling}
\end{equation}
where $\delta\mathbf{U}^{m} = \{\delta\mathbf{u}_0^{m}, \dots, \delta\mathbf{u}_{K-1}^{m}\}$ is a sequence of perturbations over the prediction horizon $K$. Each perturbation is independently sampled from a zero-mean Gaussian distribution, i.e., $\delta\mathbf{u}_k^{m} \sim \mathcal{N}(0, \Sigma)$.

For each candidate sequence $\mathbf{U}^{m}$, a corresponding state trajectory $\mathbf{X}^{m}$ is calculated forward in time (a "rollout") using the dynamics model (\ref{equ:dynamics model}) as shown in Fig.\ref{fig:three_objective}-(a). The total cost of this trajectory, $\mathcal{J}(\mathbf{U}^{m})$, is then evaluated. To keep the state and input within their dynamic limit, an external cost will be added to panelize unfeasible states. Subsequently, a weight $w_m$ is assigned to each trajectory based on its cost:
\begin{equation}
    w_m = \frac{\exp\left(-\frac{1}{\lambda} \mathcal{J}(\mathbf{U}^{m})\right)}{\sum_{i=1}^{M} \exp\left(-\frac{1}{\lambda} \mathcal{J}(\mathbf{U}^{i})\right)} \label{eq:mppi_weights}
\end{equation}
where the temperature parameter $\lambda > 0$ modulates the influence of each sample. A smaller $\lambda$ leads to a more aggressive optimization that favors only the lowest-cost trajectories.

Finally, the optimal control sequence for the current iteration, ${\mathbf{U}}^*$, is computed as the weighted average of all sampled candidate sequences as shown in Fig.\ref{fig:three_objective}-(a):
\begin{equation}
    \mathbf{U}^* = \sum_{m=1}^{M} w_m \mathbf{U}^{m} \label{eq:mppi_update}
\end{equation}
In line with the receding horizon principle, only the initial control action ${u}_0^*$ from the optimized sequence $\mathbf {U}^*$ is executed on the system. The complete sequence then serves as the new nominal trajectory $\mathbf{U}^*$ for the next control cycle.

\section{Reference-Free controller via MPPI Framework}
In this section, we begin by introducing a novel, reference-free \textit{Gate Progress} objective inspired by the reward shaping techniques used in Reinforcement Learning. Subsequently, we describe how this new objective, alongside two conventional cost functions for drone racing—namely Trajectory Tracking and Contouring Control—are integrated into this unified framework. This unified design provides the flexibility to seamlessly switch between fundamentally different control strategies.
\subsection{Gate Progress Cost}
Inspired by recent successes in reinforcement learning (RL) for autonomous racing \cite{song2023reaching}, where task-specific rewards have demonstrated state-of-the-art performance, we formulate a cost function that directly optimizes progress towards sequential gates. Unlike trajectory-based objectives, this approach eschews an explicit reference path. The cost over a prediction horizon $K$ is defined as:
\begin{equation}
\begin{aligned}
    \mathcal{J}_{gate} = \sum_{k=0}^{K-1} & \left(\|\textbf{p}_{k+1} - \textbf{p}_{gate}\|_2^2 - \|\textbf{p}_{k} - \textbf{p}_{gate}\|_2^2\right)
\end{aligned}
\label{eq:gate_progress}
\end{equation}
where $\textbf{p}_{k}$ is the quadrotor's position at prediction step $k$ and $\textbf{p}_{gate}$ is the current target gate.

A challenge arises from the objective's formulation when deploying it within the receding horizon control framework. If a predicted state passes the target gate $\textbf{p}_{\text{gate}}$ at an intermediate step $t_g$, the cost for all subsequent steps ($t > t_g$) would erroneously cause the drone to stick to the current racing gate instead of flying toward the next gate. Therefore, it is essential to dynamically switch the target gate for each state within the prediction horizon once the current gate has been passed.

\begin{figure}[h]
    \centering
    \includegraphics[width=1.0\linewidth,trim={0 95cm 75cm 0cm}, clip]{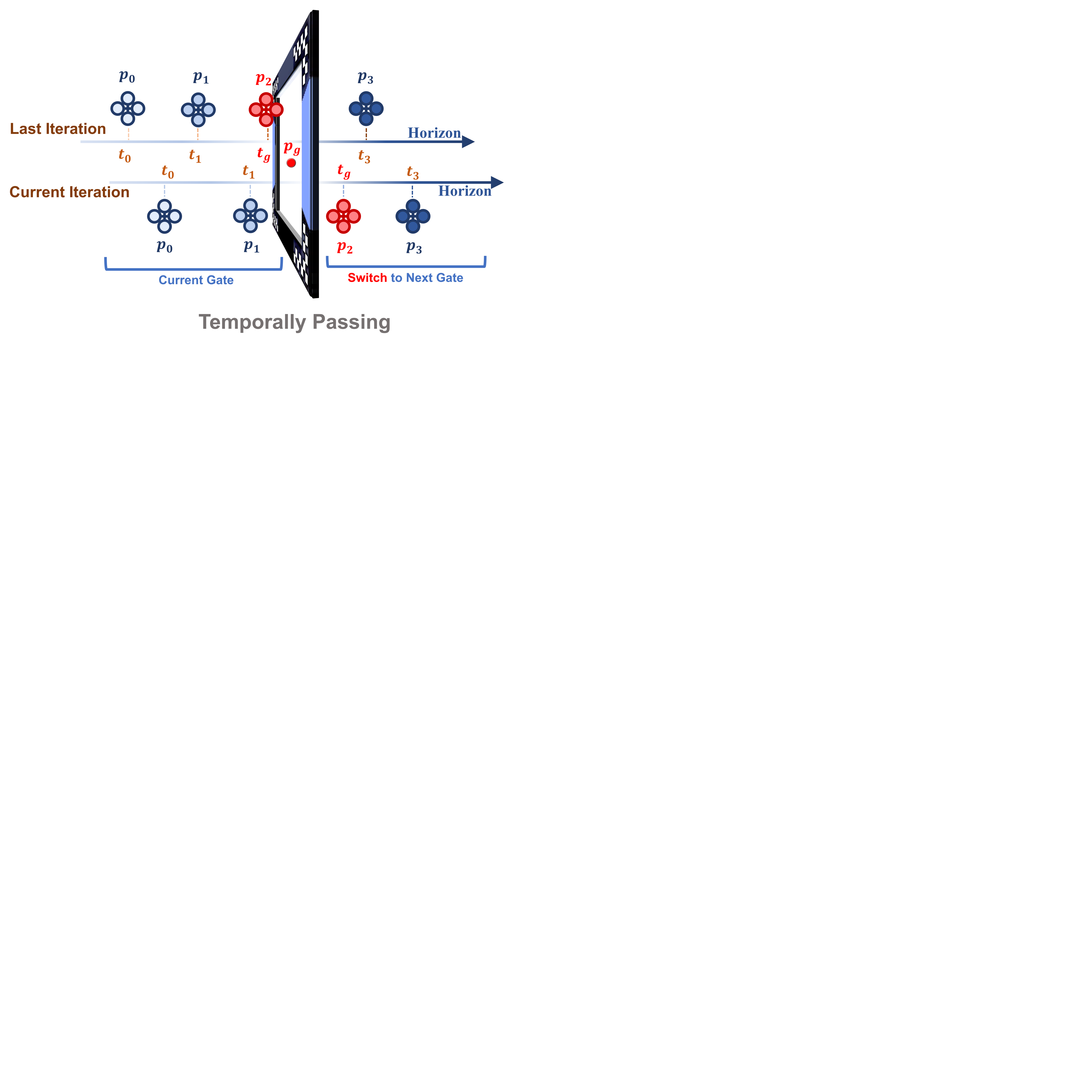}
    \caption{The position $p_2$ corresponding to time $t_2$ is behind the gate in the current iteration and in front of the gate in the previous iteration, which will be registered as a passed state $t_g$.}
    \label{fig:gate_switch}
\end{figure}

To solve this, we introduce a robust, temporally-consistent gate switching logic. While a gate pass is geometrically defined by the quadrotor flying through the gate, relying solely on this criterion can lead to chattering, where the gate-passed state might oscillate between consecutive control iterations. For instance, in the ${i-1}^{th}$ control iteration, the 3rd state may be classified as the gate-pass state, whereas in the subsequent $i^{th}$ iteration, this classification may occur at the 5th state instead, which may introduce instability. To prevent such instability, we enforce an additional temporal condition. As illustrated in Fig.~\ref{fig:gate_switch}, a gate-pass is registered in the current control iteration only if the current state satisfies the geometric crossing criteria, and the corresponding state in the \textit{previous} control iteration was determined to be in front of the gate. This combined geometric-temporal check prevents the gate-crossing point from oscillating between iterations, thereby enhancing controller stability.

With the gate switching strategy mentioned above, predicted states behind the current gate $p_{\text{gate}}^0$ are assigned to the next gate $p_{\text{gate}}^1$, as their target, as shown in the left panel of Fig.~\ref{fig:three_objective}-(b). However, when the number of states located behind the current gate becomes large, the influence of the next gate $p_{gate}^1$ becomes dominant, the progress objective (\ref{eq:gate_progress}) may adversely influence the quadrotor’s ability to successfully traverse the current gate $p_{gate}^0$. Thus, we add a penalty term to prevent states near the current gate $p_{gate}^0$(as illustrated by the red ellipse in Fig.~\ref{fig:three_objective}-(b)) from deviating too much from it. The full objective used in MPPI is:
\begin{equation}
\begin{aligned}
    \mathcal{J}_{\text{gate}} = \sum_{k=0}^{K-1} & \underbrace{\left(\|\textbf{p}_{k+1} - \textbf{p}_{\text{gate}}\|_2^2 - \|\textbf{p}_{k} - \textbf{p}_{\text{gate}}\|_2^2\right)}_{\text{gate progress}} \\
    & + Q_{\text{near}}\|\textbf{p}_{k} - \textbf{p}_{\text{gate}}\|_2^2,
\end{aligned}
\label{eq:gate_progress_aug}
\end{equation}
where $Q_{\text{near}}$ is a conditional weight that is non-zero for states near the gate determined by a predefined threshold. Using this augmented gate progress objective (\ref{eq:gate_progress_aug}), the quadrotor can fly through each gate stably and turn correctly towards the next one.
 
\subsection{A Unified Framework}

To assess the broader capabilities of our MPPI framework, we also implement two objectives that are conventionally solved with gradient-based methods. A key advantage of our unified architecture is its flexibility, which allows for these fundamentally different control objectives to be interchanged with minimal effort, such as modifying a single line of code. This enables us to not only solve a new class of problems but also to systematically and fairly compare our sampling-based approach against traditional paradigms within a single, consistent environment. In this work, we implement and systematically compare three distinct cost functions, each representing a different paradigm for autonomous drone racing. We begin with the classic \textit{Trajectory Tracking} objective, followed by the more flexible \textit{Contouring Control} method.
 
We omit the state and input range limit in the following sections for simplicity, and only keep the objective-relevant terms.

\label{sec:controllers}

\subsubsection{Trajectory Tracking Cost}
The first and most established objective we consider is trajectory tracking. This approach aligns with the classic hierarchical paradigm in robotics, where a feasible, time-parameterized reference trajectory, $\boldsymbol{\tau}_{\text{ref}}$, is first computed offline\cite{foehn2021time}. The controller's task is then to minimize the deviation from this reference online as shown in Fig.\ref{fig:three_objective}-(a). The objective is formulated as a quadratic cost function that penalizes both state tracking errors and control effort, ensuring the quadrotor follows the desired path smoothly. The cost function is defined as:
\begin{align}
    \mathcal{J}_{\text{track}} = &\sum_{k=0}^{K-1} \left( \|\mathbf{x}_k - \mathbf{x}_{k, \text{ref}}\|_{\boldsymbol{Q}}^2 + \|\mathbf{u}_k\|_{\boldsymbol{R}}^2 \right)+\|\mathbf{x}_K - \mathbf{x}_{K, \text{ref}}\|_{\boldsymbol{Q}_K}^2\label{eq:tracking_cost} \\
    \text{s.t.} \quad & \mathbf{x}_{k+1} = f(\mathbf{x}_k, \mathbf{u}_k), \quad \mathbf{x}_0 = \mathbf{x}_{\text{init}} \nonumber
\end{align}
where $\mathbf{x}_{k, \text{ref}}$ is the reference state at prediction step $k$. The positive semi-definite matrices $\boldsymbol{Q}$, $\boldsymbol{R}$, and $\boldsymbol{Q}_N$ are weighting matrices that penalize the stage state error, control input, and terminal state error, respectively.

\subsubsection{Contouring Control Cost}

Instead of tracking a time-parameterized trajectory, MPCC follows a purely geometric path $\mathcal{P}_{\text{arc}}$ parameterized by arc length, $\theta$. This grants the controller an additional degree of freedom: to dynamically optimize its speed along the path. The objective is twofold: to minimize deviation from the reference path while simultaneously maximizing the progress along it. The cost function is composed of terms for path-following error, progress maximization, and control input regularization:
\begin{equation}
\begin{aligned}
    \mathcal{J}_{contour} = \min \quad & \sum_{k=0}^{K-1} \|\boldsymbol{e}^l(\theta_k)\|_{q_l}^2 + \|\boldsymbol{e}^c(\theta_k)\|_{q_c}^2 + \|\boldsymbol{\omega}_k\|^2_{\mathbf{Q}_{\boldsymbol{\omega}}} \\
    & \qquad  + \|\Delta v_{\theta_k}\|_{r_{\Delta v}}^2 + \|\Delta\boldsymbol{T}_k\|_{\boldsymbol{R}_{\Delta\boldsymbol{T}}}^2 - \mu v_{\theta_k} \\
    \text{s.t.} \quad & \boldsymbol{x}_0 = \boldsymbol{x} \\
    & \boldsymbol{x}_{k+1} = f(\boldsymbol{x}_k, \boldsymbol{u}_k) \\
    & 0 \le v_{\theta} \le v_{\theta, \text{max}}
\end{aligned}
\label{eq:MPCC_corrected}
\end{equation}
Here, $\theta_k$ is the progress along the path, while $v_{\theta,k}$ and $\Delta v_{\theta,k}$ are its first and second time derivatives. The positional error $\boldsymbol{e}_k$ is decomposed into a lag error $\boldsymbol{e}^l_k$ (tangential to the path) and a contouring error $\boldsymbol{e}^c_k$ (perpendicular to the path) as shown in Fig.\ref{fig:three_objective}-(b). The term $-\mu v_{\theta,k}$ incentivizes maximizing forward speed. Additional terms penalize the body rates $\boldsymbol{\omega}_k$, and the rate of change of progress $\Delta v_{\theta,k}$ and thrust $\Delta\boldsymbol{T}_k$ to ensure smooth and feasible solutions. The weighting matrices and scalar $\mu$ balance these competing objectives. For more details with MPCC, we refer readers to \cite{romero2022time}.

We summarize the update loop of the MPPI framework and highlight the different control cost functions as in Algorithm \ref{alg:mppi}.
\begin{algorithm}
\caption{Model Predictive Path Integral (MPPI) Control}
\label{alg:mppi}
 \textbf{Inputs:} \\
 \hspace{\algorithmicindent} {Reference Trajectory $\tau_{ref}$}, {Arc-Parameterized Path $\mathcal{P}_{arc}$}, {Waypoints $\mathcal{W}$}\\
 \textbf{Parameters:} \\
 \hspace{\algorithmicindent} Number of samples $M$, Horizon length $K$, Temperature $\lambda$, Noise covariance $\Sigma$\\
\hrule
\begin{algorithmic}[1]
\State Initialize $\mathbf{U} \in \mathbb{R}^{M \times K} \leftarrow \mathbf{0}$
\While{task not complete}
    \State Get current state $\mathbf{x}_{0}$
    \State \textcolor{gray}{// MPPI Sampling and Evaluation Loop}
    \For{$m = 1$ to $M$} 
        \State Sample control sequence $\mathbf{U}^{m}$ using (\ref{eq:mppi_sampling})
        \State \textcolor{gray}{// Rollout state trajectory}
        \For{$k=0$ to $K-1$}
            \State $\mathbf{X}^{m}\leftarrow f(\textbf{x}_k,\textbf{u}_k^{m})$ 
        \EndFor
        \State \textcolor{gray}{// Calculate the cost using specific input}
                
        \State {$\mathcal{J}[m] \leftarrow \mathcal{J}_{\text{track}}(\mathbf{X}^{m}, \mathbf{U}^{m}, \boldsymbol{\tau}_{\text{ref}})$} or
        
        \State {$\mathcal{J}[m] \leftarrow \mathcal{J}_{\text{contour}}(\mathbf{X}^{m}, \mathbf{U}^{m}, \mathcal{P}_{\text{arc}})$} or
        
        \State {$\mathcal{J}[m] \leftarrow \mathcal{J}_{\text{gate}}(\mathbf{X}^{m}, \mathbf{U}^{m},\mathcal{W})$}
    \EndFor
    \State Compute weights using (\ref{eq:mppi_weights})
    \State Update control sequence using (\ref{eq:mppi_update})
\EndWhile
\end{algorithmic}
\end{algorithm}

\section{Experiments and Analysis}
\subsection{Simulation Results and Analysis}

Our proposed Model Predictive Path Integral (MPPI) controller was implemented using the \texttt{JAX} framework for high-performance computation based on \cite{yi2024covo}, with all simulations executed on an NVIDIA RTX 3090 GPU. The specific parameters governing the MPPI controller are detailed in Table~\ref{tab:mppi_param}. For benchmarking purposes, we compared our approach against an optimal control problem (OCP) solver based on \texttt{acados} adapted from \textit{Agilicious} \cite{doi:10.1126/scirobotics.abl6259}. This solver was selected due to its widespread adoption and established performance in the field of Model Predictive Control (MPC) for autonomous drone racing \cite{romero2022time}.

\begin{table}[h]
    \centering
     \caption{The parameters used in simulation}
    \begin{tabular}{cc|cc}
     \toprule
    \textbf{Parameters} & \textbf{Value/Range} & \textbf{Parameters} & \textbf{Value/Range} \\
    \midrule
        $M$ & $8192$ & $\lambda$ & $0.01$ \\  
        $K$ & $20$ & $\Delta t$ & $0.03$ \\
         $\Delta T_s$ & $[-10,10]$ & $\Delta v_{\theta}$ & $[-20,20]$\\ 
        $\omega[rad/s]$ & $[-10,10]$ & $T_{s}[N]$ & $[0,10]$\\ 
         \bottomrule
    \end{tabular}
     \label{tab:mppi_param}
\end{table}

To ensure a rigorous and fair comparison, a consistent quadrotor dynamics model with a unified set of parameters was employed throughout all simulation experiments. We generated a globally optimal trajectory using the CPC\cite{foehn2021time} method to serve as a common reference for two distinct control objectives. For the \textit{trajectory tracking} task, the controller was provided with the full time-parameterized state trajectory. In contrast, for the \textit{contouring control} task, only the geometric path (i.e., the position sequence) from the reference trajectory was utilized, which was subsequently converted into an arc-length parameterized representation.

\begin{figure}[t!] 
    \centering 
    \includegraphics[width=0.5\textwidth, trim={0cm 20cm 40cm 0cm}, clip]{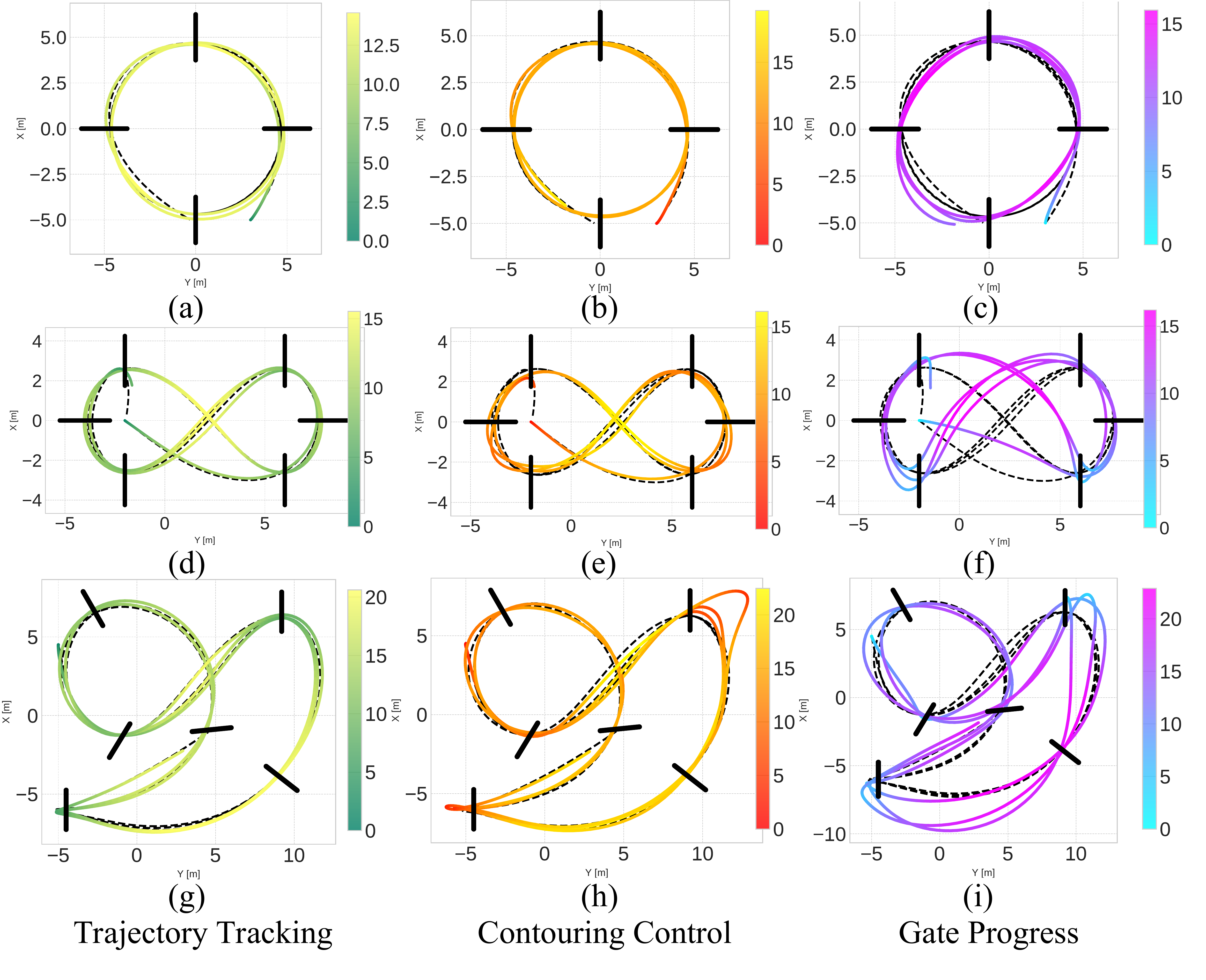}
    
    \caption{Flight trajectories generated by our unified MPPI framework on three tracks of increasing difficulty: Circle (a-c), Figure-8 (d-f), and Split-S (g-i). All the flight trajectories are tested under identical system dynamics and controller settings.}
    \label{fig:main}
    \vspace{-1em}
\end{figure}

\begin{table*}[ht]
    \centering
    \caption{Comparison across different objectives. 'Opt.' refers to the gradient-based OCP solver, while 'Ours' refers to our MPPI-based implementation.}
    \label{tab:sim_comparison}
    
    \sisetup{
        detect-weight, 
        mode=text,
        table-format=2.2 
    }
\resizebox{\textwidth}{!}{
    \begin{tabular}{
        cccccccccccccc
    }
    \toprule
    
    \multirow{3}{*}{\textbf{Track}} & \multirow{3}{*}{\textbf{CPC Reference Time (s)}} & \multicolumn{6}{c}{\textbf{Waypoint Dis (m)$\downarrow$}} & \multicolumn{4}{c}{\textbf{Total Flight Time (s) $\downarrow$}} & \multicolumn{2}{c}{\textbf{Tracking RMSE (m) $\downarrow$}} \\
    \cmidrule(lr){3-8} \cmidrule(lr){9-12}\cmidrule(lr){13-14}
    
    & &\multicolumn{2}{c}{$\boldsymbol{\mathcal{J}_\text{tracking}}$} & \multicolumn{2}{c}{$\boldsymbol{\mathcal{J}_\text{contour}}$} & \multicolumn{2}{c}{$\boldsymbol{\mathcal{J}_\text{gate}}$} & \multicolumn{2}{c}{$\boldsymbol{\mathcal{J}_\text{contour}}$} & \multicolumn{2}{c}{$\boldsymbol{\mathcal{J}_\text{gate}}$} & \multicolumn{2}{c}{$\boldsymbol{\mathcal{J}_\text{tracking}}$} \\
    \cmidrule(lr){3-4}\cmidrule(lr){5-6}\cmidrule(lr){7-8} \cmidrule(lr){9-10} \cmidrule(lr){11-12}\cmidrule(lr){13-14}

    & & {Opt.} & {\textbf{Ours}} & {Opt.} & {\textbf{Ours}} & {Opt.} &{\textbf{Ours}} & {Opt.} & {\textbf{Ours}} & {Opt.} &{\textbf{Ours}} & {Opt.} & {\textbf{Ours}} \\
    \midrule

    {Circle} 
    & \cellcolor{gray!15} 7.29 & \cellcolor{gray!15} 0.382 & \cellcolor{gray!15} 0.327 & \cellcolor{gray!15} 0.484 & \cellcolor{gray!15} 0.367 & \cellcolor{gray!15} -& \cellcolor{gray!15} 0.429 & \cellcolor{gray!15} 7.65 & \cellcolor{gray!15} 7.14 & \cellcolor{gray!15} -& \cellcolor{gray!15}{7.47} & \cellcolor{gray!15} 0.25 ± 0.14 & \cellcolor{gray!15} 0.32 ± 0.23 \\

    {Figure-8} 
    &  10.72 & 0.486 &0.424 &0.595 &0.493 & - &0.380 &  12.03 &  10.26 & - & 11.16 &  0.23 ± 0.13 & {0.14 ± 0.07} \\

    {Split-S} 
    & \cellcolor{gray!15} 16.52 & \cellcolor{gray!15} 0.491 &\cellcolor{gray!15} 0.354 &\cellcolor{gray!15} 0.69 &\cellcolor{gray!15} 0.375 & \cellcolor{gray!15} -&\cellcolor{gray!15} 0.527 & \cellcolor{gray!15} 17.10 & \cellcolor{gray!15} 17.43 & \cellcolor{gray!15} -& \cellcolor{gray!15}{17.16} & \cellcolor{gray!15} 0.39 ± 0.23 & \cellcolor{gray!15} {0.24 ± 0.14} \\
    \bottomrule
    \end{tabular}
    }
    \vspace{-1em}
\end{table*}

We tested the proposed framework on three tracks of increasing difficulty: \textit{Circle}, \textit{Figure-8}, and \textit{Split-S}. For each track, the quadrotor needs to fly through a sequence of gates for three laps. We choose the total flight time to evaluate the performance of the contouring objective function and tracking RMSE for the trajectory tracking objective function as in Table \ref{tab:sim_comparison}. We also use the waypoint pass error, which is defined as the state that has the minimal distance to the waypoint, to indicate the overall performance in a racing track.

The quantitative results of our comparative study are summarized in Table~\ref{tab:sim_comparison}. We evaluated the performance of our proposed reference-free \textit{gate progress} objective ($\mathcal{J}_\text{gate}$). The results indicate that this control objective, which does not rely on any reference trajectory, can achieve near time-optimal flight performance. Its completion times are highly competitive with the specialized contouring controller, and it successfully maintains a low waypoint traversal error, highlighting its efficacy as a powerful alternative for time-sensitive tasks.

Furthermore, for the trajectory tracking task ($\mathcal{J}_\text{tracking}$), our MPPI-based implementation demonstrates superior performance by consistently reducing the waypoint traversal distance and achieving a lower Tracking Root Mean Square Error (RMSE) compared to the optimal baseline. In the context of the contouring control task ($\mathcal{J}_\text{contour}$), our method yields shorter total flight times on the Circle and Figure-8 tracks while maintaining a lower waypoint distance across all three courses.

\begin{figure}[h]
    \centering
    \includegraphics[width=1.0\linewidth,trim={2cm 0cm 3cm 1cm}, clip]{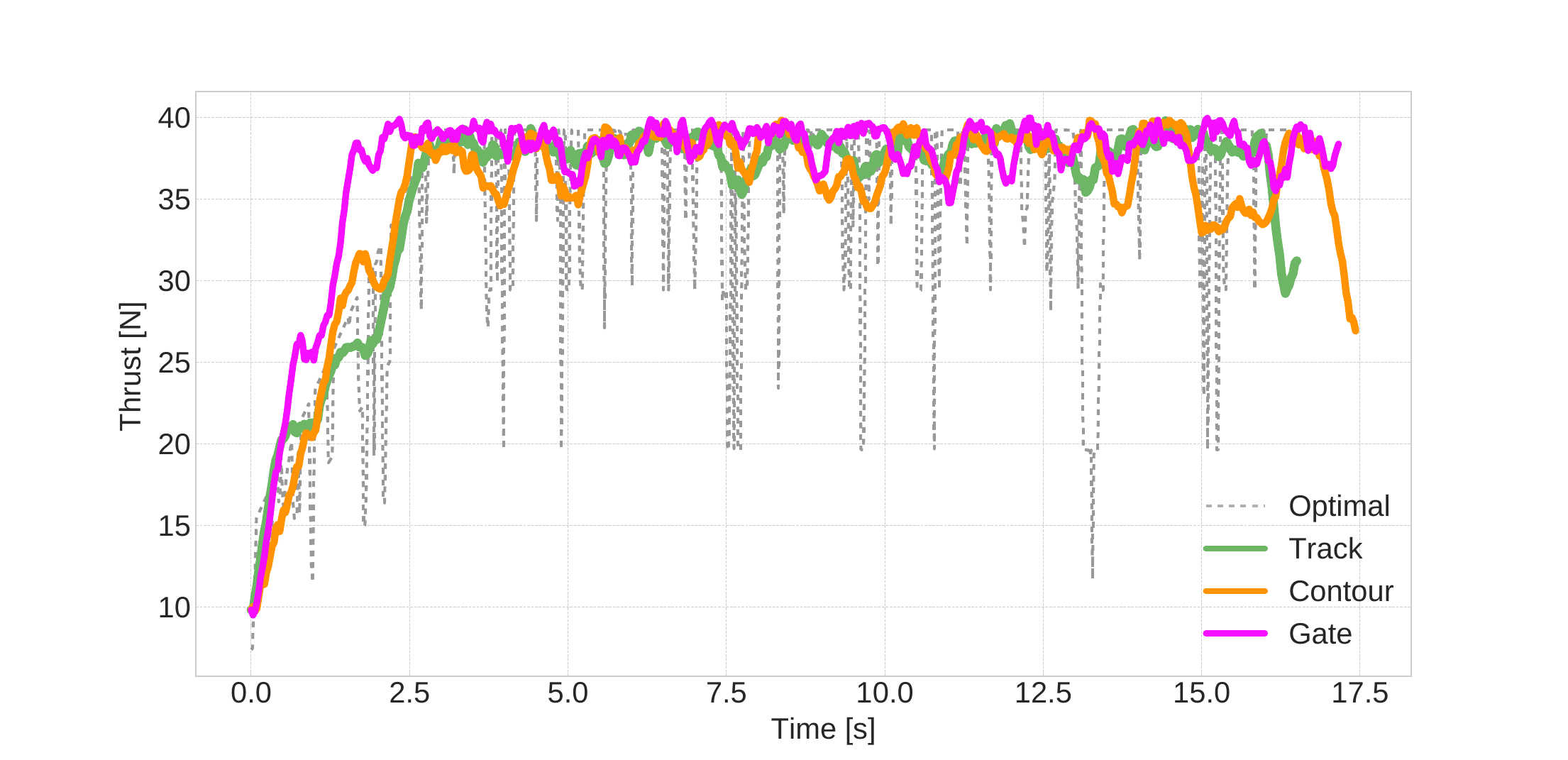}
    \caption{Comparison of the total commanded thrust profiles for each control objective in the Split-S track.}
    \label{fig:thrust}
    \vspace{-1em}
\end{figure}

\begin{figure*}[h]
    \centering
    \includegraphics[width=1.0\linewidth,trim={0cm 100cm 10cm 0cm}, clip]{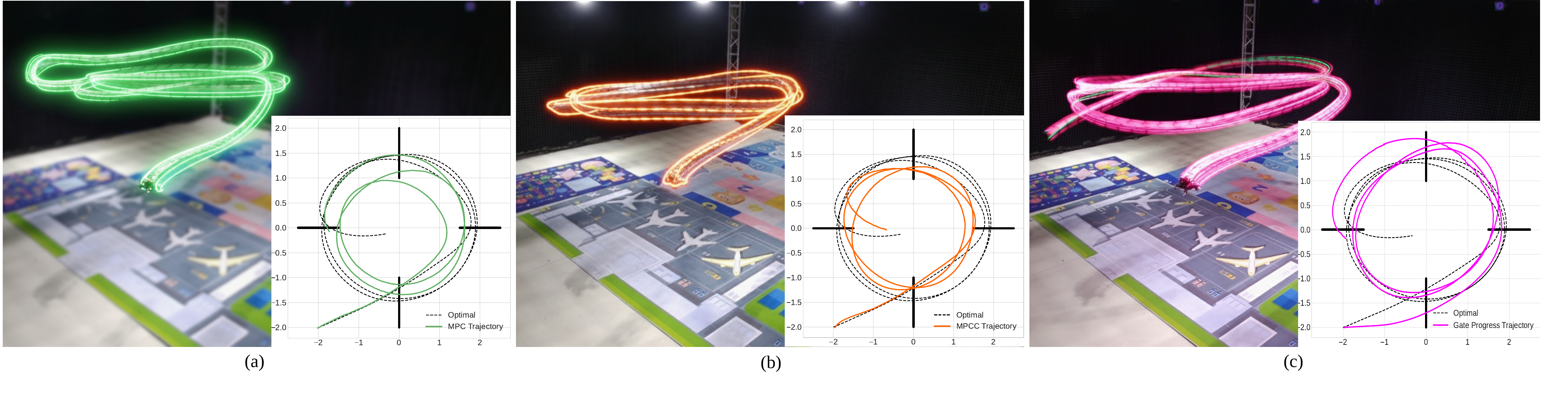}
    \caption{Real-world flight validation of the three control objectives executed within our unified MPPI framework. The panels depict the quadrotor's performance using: (left) Trajectory Tracking, (middle) Contouring Control, and (right) the proposed reference-free Gate Progress objective. All three strategies were deployed using identical dynamics parameters and a consistent control input structure.}
    \label{fig:realflight}
    \vspace{-1em}
\end{figure*}

To provide a more intuitive understanding of the controllers' behavior in the time-optimal task, we visualize the total thrust command for each strategy, as shown in Fig.~\ref{fig:thrust}. The trajectory tracking controller exhibits a relatively conservative thrust profile. In contrast, the contouring controller employs a more aggressive strategy, frequently pushing the system towards its input limits to minimize lap time. Most notably, the controller optimizing the \textit{gate progress} objective drives the system to its maximum thrust limits earliest and sustains this input saturation for the longest duration. This aggressive behavior corroborates the quantitative results and visually confirms that the \textit{gate progress} objective most directly and effectively prioritizes the minimization of total flight time.

\subsection{Real‐world Experiments}

For the real-world validation, we employed a customized quadrotor with a total weight of $340\,\text{g}$, which has a thrust-to-weight ratio $(\text{TWR})=3$. All onboard computation was performed by an embedded Jetson Orin NX (16GB) module. The experiments were conducted on a small, circular track designed specifically to benchmark the performance of our proposed MPPI-based method against a conventional optimization-based controller in a physical environment. This setup allows for a direct comparison of their practical applicability and real-world flight characteristics.

\begin{table}[ht]
    \centering
    \caption{Solving Time on Different Platforms}
    \label{tab:computation_time_comparison}

    \sisetup{
        detect-weight,
        table-format=3.2
    }
    \begin{tabular}{ccc}
    \toprule
    \multirow{2}{*}{\textbf{Objective/Method}} & \multicolumn{2}{c}{\textbf{Solving Time (ms) $\downarrow$}} \\
    \cmidrule(lr){2-3}
    & {\textbf{Desktop}} & {\textbf{Jetson NX}} \\
    
    \midrule
    Tracking/Opt. & 0.8 & \textbf{3.0} \\
    Contouring/Opt. & 1.4 & 4.0 \\
    \midrule
    MPPI & $\textbf{0.4}$ & 6.7 \\

    \bottomrule
    \end{tabular}
    \vspace{-1em}
\end{table}

First, the computational performance of the MPPI and OCP methods was benchmarked on a desktop computer with an RTX 3090 GPU and an embedded Jetson NX. A summary of the solving times is presented in Table~\ref{tab:computation_time_comparison}, and the memory used by the MPPI is about 500MB. On the desktop platform, the MPPI controller demonstrates exceptional efficiency. Leveraging parallel computation and just-in-time (\texttt{jit}) compilation, it achieves a consistent solving time of approximately $0.4\,\text{ms}$, which does not significantly change with the choice of objective function. In contrast, the solving time for the OCP method varies from $0.8\,\text{ms}$ to $1.4\,\text{ms}$ depending on the selected objective.
When deployed on the resource-constrained Jetson Orin NX, a notable increase in computation time is observed for both methods due to power limitations and lower memory bandwidth. The MPPI solving time increases to approximately $6.7\,\text{ms}$, while the OCP solver's time rises to a range of $3\,\text{ms}$ to $4\,\text{ms}$. This highlights a critical trade-off: while MPPI offers greater flexibility in objective function design, its computational demands on current-generation embedded hardware are higher than those of highly optimized gradient-based solvers. These findings confirm the feasibility of both methods for real-time onboard implementation. To meet the real-time requirement of a 50~Hz control frequency during the real-world flights, we set the number of samples to $M=2048$.

\begin{table}[ht]
    \centering
    \caption{Comparison across different objectives in real-world flight.}
    \label{tab:real_comparison}
    \sisetup{
        detect-weight,
        mode=text,
        table-format=2.2
    }
    \resizebox{0.48\textwidth}{!}{
    \begin{tabular}{
        cccccc
    }
    \toprule
    
     \multirow{3}{*}{\textbf{CPC Time (s)}} & \multicolumn{3}{c}{\textbf{Flight Time (s) $\downarrow$}} & \multicolumn{2}{c}{\textbf{Tracking RMSE (m) $\downarrow$}} \\
    \cmidrule(lr){2-4} \cmidrule(lr){5-6}
    
    &  \multicolumn{2}{c}{$\boldsymbol{\mathcal{J}_\text{contour}}$} & {$\boldsymbol{\mathcal{J}_\text{gate}}$} & \multicolumn{2}{c}{$\boldsymbol{\mathcal{J}_\text{tracking}}$} \\
    \cmidrule(lr){2-3}\cmidrule(lr){4-4}\cmidrule(lr){5-6}

    & {Opt.} & {\textbf{Ours}} & {\textbf{Ours}} & {Opt.} & {\textbf{Ours}} \\
    \midrule

    6.33 & 5.84 & 7.13 & 6.56 & 0.36 ± 0.14 & 0.46 ± 0.18\\
    
    \bottomrule
    \end{tabular}
    }
    \vspace{-1em}
\end{table}

We list the key metrics from our real-world experiments in Table \ref{tab:real_comparison}, and long-exposure photographs capturing the flight trajectories are shown in Fig.~\ref{fig:realflight}. As the results indicate, our proposed reference-free gate progress objective achieved a flight time of 6.56s, demonstrating near time-optimal performance in a real-world setting when compared to the theoretical optimal time of 6.33s. Furthermore, it is noteworthy that the gate progress method achieves a lower gate-passing error compared to its reference-based counterparts. This is directly attributable to its objective function, which is explicitly formulated to optimize for the primary racing task of gate traversal. In contrast, the other two methods focus on the surrogate goal of tracking a reference path. This superior performance in passing through the gates empirically demonstrates that the gate progress objective is more task-relevant. For a rigorous benchmark of the traditional objectives, we solved the same Optimal Control Problems (OCPs) for trajectory tracking and contouring control using the gradient-based solver \texttt{acados}, while keeping all controller parameters strictly identical. The comparison reveals a performance trade-off for our MPPI implementation on the embedded hardware. The tracking RMSE is higher (0.46 vs. 0.36), and the flight time for the contouring controller is slower (7.13s vs. 5.84s). We attribute this performance gap to the reduced number of samples (M=2048) necessitated by the computational constraints of the onboard computer. This highlights a key insight: the performance of sampling-based methods is intrinsically linked to computational budget. 

\section{Conclusion}
We have introduced a reference-free control strategy for agile drone racing that achieves near time-optimal performance by incorporating a reinforcement learning-inspired "gate progress" objective into a model-based controller. Our unified framework, built upon the Model Predictive Path Integral (MPPI) control objective, uniquely handles this non-differentiable objective while also providing a fair and flexible testbed for traditional trajectory tracking and contouring control, with seamless switching between methods. This work offers a valuable platform for systematic comparison between diverse control objectives.

While our approach excels on highly parallel hardware like desktop GPUs, we acknowledge that this advantage does not fully translate to current embedded systems like the Jetson Orin NX due to hardware constraints such as memory bandwidth. However, we contend that the trend toward more powerful parallel processors will continue to validate the utility of sampling-based methods for complex, real-world robotics tasks. Future efforts will be directed toward designing more efficient frameworks to lower the computational barrier and exploring more advanced objective functions within the sampling-based paradigm to further enhance flight agility and autonomy.

\bibliography{reference}
\bibliographystyle{ieeetr}
\end{document}